\def\year{2019}\relax
\documentclass[letterpaper]{article} %DO NOT CHANGE THIS
\usepackage{aaai19}  %Required
\usepackage{times}  %Required
\usepackage{helvet}  %Required
\usepackage{courier}  %Required
\usepackage{url}  %Required
\usepackage{graphicx}  %Required
\usepackage{paralist}  %added
\usepackage{xspace}   %added
\usepackage{amsmath}
\usepackage{multirow}
\usepackage{color}
\newcommand{\method}{CGMH\xspace}
\title{CGMH: Constrained Sentence Generation by Metropolis-Hastings Sampling}
\author{Ning Miao\textsuperscript{1}\thanks{Work done while Ning Miao was a research intern in ByteDance AI Lab. Hao Zhou and Rui Yan are corresponding authors.}\quad Hao Zhou\textsuperscript{2}\quad Lili Mou\textsuperscript{3}\quad Rui Yan\textsuperscript{1}\quad Lei Li\textsuperscript{2}\\
\textsuperscript{1}Institute of Computer Science and Technology, Peking University, China
\\
\textsuperscript{2}ByteDance AI Lab, Beijing, China\quad
\textsuperscript{3}AdeptMind Research, Toronto, Canada\\
miaoning@pku.edu.cn, zhouhao.nlp@bytedance.com\\
lili@adeptmind.ai, doublepower.mou@gmail.com\\
ruiyan@pku.edu.cn,
lileilab@bytedance.com}
\newcommand{\newcite}[1]{\citeauthor{#1}~\shortcite{#1}}
\begin{document}

\maketitle
\begin{abstract}
 
In real-world applications of natural language generation, there are often constraints on the target sentences in addition to fluency and naturalness requirements. Existing language generation techniques are usually based on recurrent neural networks~(RNNs). 
However, it is non-trivial to impose constraints on RNNs while maintaining generation quality, 
since RNNs generate sentences sequentially (or with beam search) from the first word to the last. In this paper, we propose \method, a novel approach using Metropolis-Hastings sampling for constrained sentence generation. \method allows complicated constraints such as the occurrence of multiple keywords in the target sentences, which cannot be handled in traditional RNN-based approaches. Moreover, \method works in the inference stage, and does not require parallel corpora for training. We evaluate our method on a variety of tasks, including keywords-to-sentence generation, unsupervised sentence paraphrasing, and unsupervised sentence error correction. 
\method achieves high performance compared with previous supervised methods for sentence generation. Our code is released at https://github.com/NingMiao/CGMH
\end{abstract}

\section{Introduction}

\begin{figure}[!t]
   \centering
   \includegraphics[width=.47\textwidth]{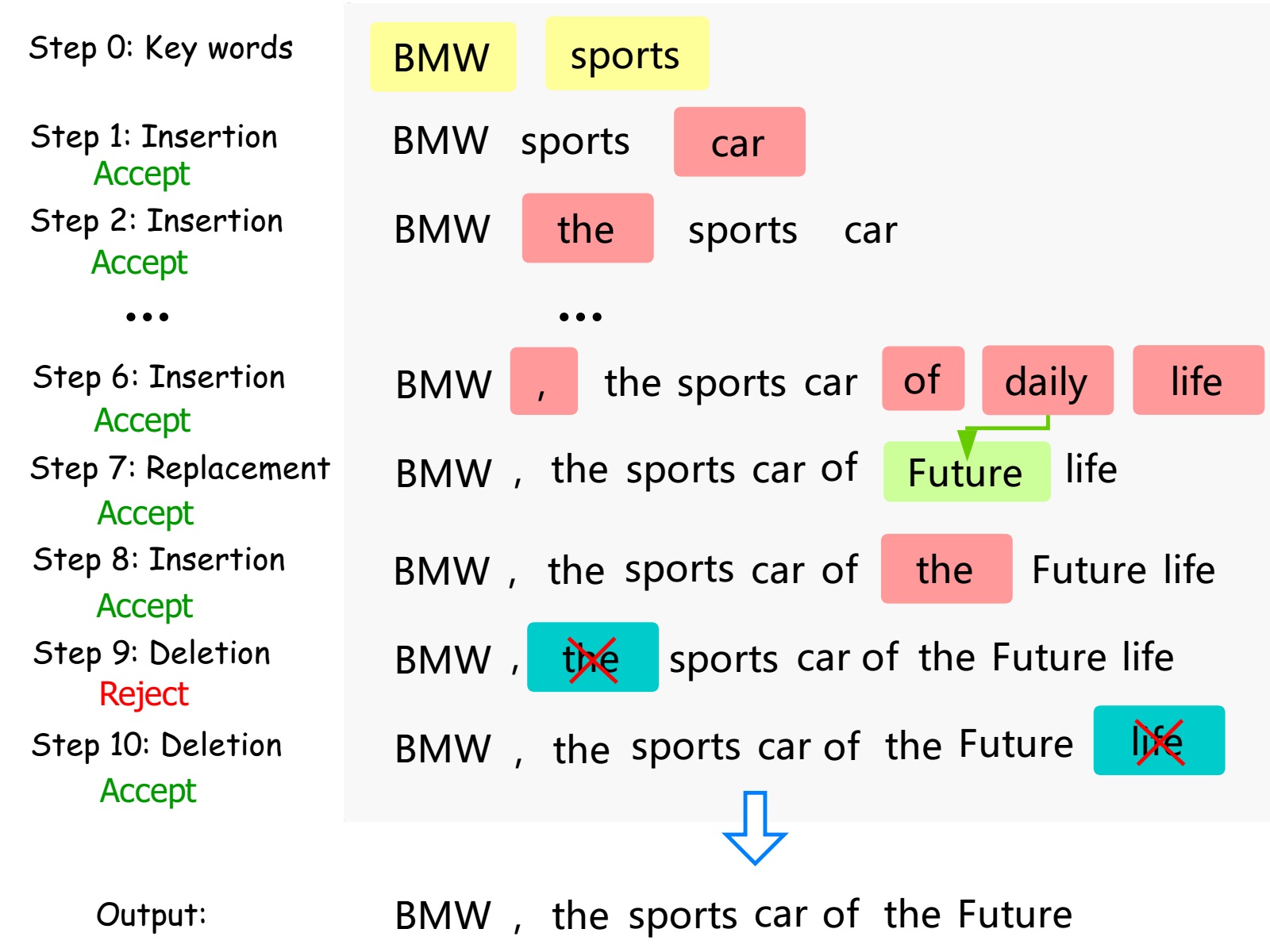}
    \caption{\method generates a sentence with the constraint of keyword inclusion. At each step, \method proposes a candidate modification of the sentence, which is accepted or rejected according to a certain acceptance rate.}
    \label{fig:intro_example}
\end{figure}

Natural language generation oftentimes involves constraints on the generated sentences. The constraints can be categorized into the following types:
\begin{inparaenum}[(1)]
\item Hard constraints, such as the mandatory inclusion of certain keywords in the output sentences; and
\item Soft constraints, such as requiring the generated sentences to be semantically related to a certain topic. 
\end{inparaenum} 
Figure~\ref{fig:intro_example} illustrates an example of advertisement generation, where ``\textit{BMW}'' and ``\textit{sports}'' should appear in the advertising slogan. Hence, ``\textit{BMW, the sports car of the future}'' is a valid sentence as an advertisement. 

Existing sentence generation methods are mostly based on recurrent neural networks (RNNs), which generate a sentence sequentially from left to right~\cite{sutskever2014sequence}.
However, it is non-trivial to impose constraints during the left-to-right generation in RNNs. 
Previous work proposes a backward-forward generation approach~\cite{mou2015backward}, which could only generate sentences with one keyword. 
Additionally, researchers propose grid beam search to incorporate constraints in machine translation~\cite{post2018fast,hasler2018neural,hokamp2017lexically}. 
It works with the translation task because the source and target are mostly aligned and the candidate set of translations is small. 
But grid beam search would not work well with general sentence generation, which has many more candidate sentences.

In this paper, we propose \textit{Constrained Generation by Metropolis-Hastings sampling} (\method), a novel approach that addresses sentence generation with constraints. Different from previous sentence samplers working in the variational latent space~\cite{K16-1002},
\method directly samples from the sentence space using the Metropolis-Hastings (MH) algorithm~\cite{metropolis1953equation}. 
MH is an instance of the general Markov chain Monte Carlo (MCMC) sampling. It defines local operations in the sentence space (e.g., word replacement, deletion, and insertion). 
During sampling, we randomly choose a word and an {operation} to form a \textit{proposal} for transition. 
The proposal is either accepted or rejected according to an \textit{acceptance rate} computed by a pre-specified \textit{stationary distribution}. 
Compared with Gibbs sampling (another MCMC method), MH is more flexible to generate sentences with arbitrary lengths.

%\method naturally generates more diverse samples compared with RNN-based generation algorithms. 

It is then straightforward to impose constraints on the generated sentences by introducing a matching function (which indicates the degree to which the constraints are satisfied) to manipulate the stationary distribution of MH. For hard constraints, the matching function could be a binary indicator, ruling out the possibility of an infeasible sentence. For soft constraints, the matching function could be, for example, a measure of semantic similarity. In both cases, we are able to sample sentences with constraints.

Our proposed \method can be applied to a variety of tasks. 
We first experiment keywords-to-sentence generation, where keywords are hard constraints. In this task, \method outperforms state-of-the-art constrained generation models in both negative likelihood~(fluency) and human evaluation.
We also conduct experiments on two generation tasks with soft constraints, namely, paraphrase generation and sentence error correction. Results show that, without a parallel corpus, \method significantly outperforms other unsupervised models, achieving close results to state-of-the-art supervised approaches.

In summary, our contributions include
\begin{compactitem}
\item We propose \method, a general framework for sentence sampling that can cope with both hard and soft constraints. 
\item We design the proposal distribution and stationary distributions for MH sentence sampling in three tasks, including keywords-to-sentence generation, paraphrase generation, and sentence error correction. Experimental results show that \method achieves high performance compared with previous methods.
\item We make it possible to accomplish the above tasks in an unsupervised fashion, which does not require a parallel corpus as is needed in previous approaches.
\end{compactitem}

% Sentence correction is an example of soft constraints, where the meaning of original sentences should not be changed.
% 3) Combination of 1) and 2). Paraphrasing can be cast into the third case. In paraphrase, a rewritten sentence should represent the same semantic meaning as the original utterance, though with different surface terms. At the same time, some important keywords must not be removed during paraphrasing.

% Variational autoencoders (VAEs) extends RNN, adding prior to the distribution of latent variable~\citeauthor{kingma2013auto}. However, the training of VAE is difficult because the posterior distribution of the latent variable may collapse of get uninformative. Also, VAE suffers from the problem of error accumulation, similar to basic autoencoder (AE), as their decoders share the same structure. 

\section{Related Work}

In recent years, sentence generation is mostly based on the recurrent neural network (RNN) because of its capability of learning highly complicated structures of language. In most tasks, RNN-based sentence generation is modeled as max \textit{a posteriori} (MAP) inference, and people use greedy search or beam search to approximate the most probable sentences~\cite{sutskever2014sequence}.

For sentence sampling, the most na\"ive approach, perhaps, is to sample words from RNN's predicted probabilities step-by-step, known as \textit{forward sampling} in the Bayesian network regime. The prototype-then-edit model~\cite{guu2018generating} first samples prototypes (sentences in the training set), and then edits the prototypes to obtain new sentences; it can be thought of as sampling from an RNN-defined kernel density. 
\newcite{K16-1002} use variational autoencoders (VAEs) to sample sentences from a continuous latent space. However, these approaches allow neither soft constraints that require flexible manipulation of sentence probabilities, nor hard constraints that specify one or more given words.

\newcite{berglund2015bidirectional} propose a Gibbs sampling model that uses a bi-directional RNN to alternatively replace a token from its posterior distribution. \newcite{su2018incorporating} further apply it to control the sentiment of a sentence. However, the shortcoming of Gibbs sampling is obvious: it cannot change the length of sentences and hence is not able to solve complicated problems such as sentence generation from keywords. Our paper extends Gibbs sampling with word insertion and deletion. This results in a Metropolis-Hastings sampler, enabling more flexible sampling.
\newcite{AIIDE1715908} utilize MH to sample an event sequence in the task of story generation, which cannot be directly used for sentence generation.

Another line of work tackles the problem of constrained sentence generation from a searching perspective. In neural machine translation, for example, grid beam search~\cite[GBS]{hokamp2017lexically} makes use of 2-dimensional beam search to seek sentences that satisfy the constraints, whereas constrained beam search~\cite[CBS]{anderson2017guided} utilizes a finite-state machine to assist searching. \newcite{post2018fast} and \newcite{hasler2018neural} further accelerate the search process. In machine translation, the search space is limited and highly conditioned on the source sentence. But in other generation tasks, there may be many more candidate sentences; GBS fails when the greedy pruning is unable to find sentences satisfying the constraints due to the large search space.

Sentence generation with soft constrains is also related to controlling latent features of a sentence~\cite{P18-1080,N18-1169}, such as the meaning and sentiment. \newcite{hu2017toward} apply a discriminator to VAE to generate sentences with specified sentiments, and \newcite{shen2017style} achieve style transfer by cross-alignment with only non-parallel data. Such approaches require an explicit definition of the latent feature (e.g., sentiment), supported by large labeled datasets.

The difference between our model and previous work is that our MH sampling framework is applicable to both hard and soft constraints. It immediately enables several non-trivial applications, including unsupervised paraphrase generation, unsupervised error correction, and keyword-based sentence generation. 

\section{Approach}
\label{sec:approach}

In this section, we describe our CGMH model (referring to \textit{Constrained Generation by Metropolis-Hastings}) in detail. We first introduce the general framework of MH sampling, and then we design MH components---including proposal design, stationary distributions, and acceptance decision---in the scenario of constrained sentence generation.

\subsection{The Metropolis-Hastings Algorithm}
The Metropolis-Hastings (MH) algorithm is a classical Markov chain Monte Carlo (MCMC) sampling approach. MCMC generates a sequence of correlated samples by iteratively jumping from one state to another, according to the transition matrix of a Markov chain.  For sentence generation, each state of the Markov chain is a particular sentence. Under mild conditions, the distribution of samples will converge to the \textit{stationary distribution} of the Markov chain. Therefore, we need to design a Markov chain whose stationary distribution is the desired sentence distribution. 

However, it is sometimes difficult to directly specify the transition probability of the Markov chain to obtain an arbitrary stationary distribution. The MH sampler solves this problem in a two-step fashion. It first proposes a tentative transition, but then accepts or rejects the \textit{proposal} according an \textit{acceptance rate}. The acceptance/rejection rate is computed by the desired stationary distribution and the proposal distribution. This ensures the \textit{detailed balance} condition, which in turn guarantees that MH converges to the desired distribution.

More specifically, let $\pi(\mathrm{x})$ be the distribution from which we want to sample  sentences ($\mathrm{x}$ denotes a particular sentence). 
MH starts from a (possibly) arbitrary state $\mathrm{x}_0$~(an initial sentence or a sequence of keywords). 
At each step~$t$, a new sentence $\mathrm{x}'$ is proposed based on a proposal distribution ${g}(\mathrm{x}'|\mathrm{x}_{t-1})$, where $\mathrm{x}_{t-1}$ denotes the sentence of the last step. 
Then the proposal could be either accepted or rejected,  given by the acceptance rate
\begin{align}\label{eqn:acceptance1}
A (\mathrm{x}'|\mathrm{x}_{t-1})&=\min\{1, A^*(\mathrm{x}'|\mathrm{x}_{t-1})\} \\
\label{eqn:acceptance2}
\text{where}\quad\quad A^*(\mathrm{x}'|\mathrm{x}_{t-1})&=\frac{\pi(\mathrm{x}'){g}(\mathrm{x_{t-1}}|\mathrm{x}')}{\pi(\mathrm{x}_{t-1}){g}(\mathrm{x}'|\mathrm{x}_{t-1})}
\end{align}

In other words, the proposal is accepted with a probability of $A(\mathrm x'|\mathrm{x}_{t-1})$, and the next sentence $\mathrm x_t=\mathrm x'$. 
With a probability of $1-A(\mathrm x'|\mathrm{x}_{t-1})$, the proposal is otherwise rejected and $\mathrm x_t=\mathrm x_{t-1}$.
Theoretically, the distribution of sample $\mathrm{x}_n$ will converge to $\pi(\mathrm x)$ as $n\rightarrow \infty$ for irreducible and aperiodic Markov chains. In practice, initial several samples are discarded as they are subject to the initial states $\mathrm x_0$. If the samples converge to the stationary distribution, we say the Markov chain \textit{mixes} or \textit{burns in}.
Readers may refer to \newcite{BDA} for details of MH sampling.

The MH framework is a flexible sampling algorithm because: (1) The proposal distribution could be arbitrary, as long as the Markov chain is irreducible and aperiodic; (2) The stationary distribution could also be arbitrarily specified, which will be reflected in Equation~\ref{eqn:acceptance2} to correct the proposal distribution; and (3) We can safely ignore a normalization factor of the stationary distribution and only specify an unnormalized measure, because $\pi(\cdot)$ appears in both numerator and denominator of Equation~\ref{eqn:acceptance2}. All these allow flexible manipulation of the stationary distribution.

The design of proposal distributions and stationary distributions relies heavily on applications, which will be described in the rest of this section.

\subsection{Proposals}

We design a set of simple yet effective proposals, including word \textit{replacement}, \textit{insertion}, and \textit{deletion}. 
That is, we randomly select a word at each step, and for the selected word, we randomly perform one of the three operations with probability  $[p_\text{insert}, p_\text{delete}, p_\text{replace}]$, which is set to $[1/3, 1/3, 1/3]$ in our experiments.
We further describe the operations as follows.

\paragraph{Replacement.} Assume that the sentence at the current step is $\mathrm x  = [w_1,\cdots,w_{m-1},w_m,w_{m+1},\cdots,w_n]$, where $n$ is sentence length, and that we decide to propose a replacement action for the $m$th word $w_m$.

Given all other words in the current sentence, we need to choose a new word for the $m$-th position by the conditional probability: 
\begin{equation}
\begin{aligned}
\label{replace}
&g_{\text{replace}}(\mathrm x'|\mathrm x)=\pi(w_m^*=w^c|\mathrm x_{-m}) =\\
&\frac{\pi(w_1,\cdots,w_{m-1},w^c,w_{m+1},\cdots,w_n)}{\sum_{w \in \mathcal{V}} \pi(w_1,\cdots,w_{m-1},w,w_{m+1},\cdots,w_n)}
\end{aligned}
\end{equation}
where $w_m^*$ is the new word for position $m$, $w^c$ is a candidate word for $w_m^*$, $\mathrm x'  = [w_1,\cdots,w_{m-1},w^c,$  $w_{m+1},\cdots,w_n]$ is the candidate sentence, $\mathcal{V}$ is the set of all words, and $g_{\text{replace}}(\mathrm x'|\mathrm x)$ is the probability of choosing $\mathrm x'$ as the target of replacement action from $\mathrm x$.
However, it is difficult to compute $\pi (w_m^*=w^c|\mathrm x_{-m})$ for all $w^c \in \mathcal{V}$, because we have to compute $\pi (w_1,\cdots,w_{m-1},w^c,w_{m+1},\cdots,w_n)$ for each candidate sentence separately. 
This results from different words in the middle of a sentence and thereafter different RNN hidden states.

We propose to pre-select a subset of plausible candidate words. It is easy to compute $\pi (w_1,...,w_{m-1},w_m^*=w^c)$ as well as $\pi (w_m^*=w^c,w_{m+1},...,w_n)$ by a forward and a backward language model, and $\pi (w_1,...,w_m^*=w^c,x_{n})$ is no greater than either of them.
We thus build a pre-selector $Q$ to discard $w^c$ with low forward or backward probability:
\begin{align}
\label{replacepre}
\nonumber Q(w^c)=\min(\: &\pi (w_1,...,w_{m-1},w_m^*=w^c),\\
&\pi (w_m^*=w^c,w_{m+1},...,w_n))
\end{align}

After pre-selection, we compute the conditional probability of selected words by Equation~(\ref{replace}), from which we sample a word for replacement.

\paragraph{Insertion and deletion.}
Inserting a word is done in a two-step fashion: we first insert a special token, placeholder $<$PHD$>$, at the position that we are currently working on, and then use (\ref{replace}) to sample a real word to replace the placeholder. As a result, $g_{\text{insert}}$ takes a similar form to (\ref{replace}).

Deleting is, perhaps, the simplest operation, and we directly remove the current word. Suppose $\mathrm x = [w_1, \cdots, w_{m-1},  w_m, w_{m+1} \cdots w_n]$ and we are about to delete the word $w_m$. Then
$g_{\text{delete}}(\mathrm x'|\mathrm x_{t-1})$ equals 1 if $\mathrm x' = [w_1, \cdots, w_{m-1}, w_{m+1} \cdots w_n]$, or 0 for other sentences.

Insertion and deletion ensure the \textbf{ergodicity} of the Markov chain, as in the worst case, any two sentences, $\mathrm x$ and $\mathrm x'$, are still reachable by first deleting all words in $\mathrm x$, and then inserting all words in $\mathrm x'$ in order. In addition, word replacement is an intuitive operation that helps reach ``semantically neighboring'' states more easily. 
Therefore we include it as one of our proposals. 
It should be noticed that, although the replacement action is restricted to top-ranked candidate words for efficiency purposes, this does not affect the ergodicity of the Markov chain. 

\subsection{Stationary Distribution}
\label{sec:stationary}

In the proposed CGMH framework, we would like to obtain sentences from a desired distribution $\pi(\mathrm x)$, known as the stationary distribution of the Markov chain. For constrained sentence generation, CGMH allows flexible design of the stationary distribution.

Generally, the stationary distribution $\pi(\mathrm x)$ can be defined as\\[-.7cm]
\begin{align}
\label{equ:paraphrase_stationary_general}
\pi(\mathrm x) \propto p(\mathrm x)\cdot \underbrace{\mathcal{X}_{\text{c}}^{0}(\mathrm x)\cdots \mathcal{X}_{\text{c}}^{n}(\mathrm x) }_{\text{constraints}}
\end{align}
where $p(\mathrm x)$ is the probability of a sentence in general, and $\mathcal{X}_{\text{c}}^{0}(\mathrm x),\cdots, \mathcal{X}_{\text{c}}^{n}(\mathrm x) $ are scoring functions indicating the degree to which a constraint is satisfied.
Technically, CGMH works with both hard and soft constraints.
For a hard constraint, $\mathcal{X}_{\text{c}}^{i}$ is an indicator function, which equals $1$ if the $i$th  constraint is satisfied, or 0 otherwise.
For a soft constraint, $\mathcal{X}_{\text{c}}^{i}$ is a ``smoothed'' indicator function showing how the sentence satisfies the (soft) constraint. 
  By multiplying these scoring functions together, we could impose more than one constraints.

The design of the scoring functions is flexible but task related. In our paper, we apply the CGMH framework to three different tasks.

\paragraph{Sentence generation with keywords.} In this task, we would like to generate a sentence given one or more keywords as constraints.
It has been previously explored in various applications including question answering~\cite{QA} and dialog systems~\cite{seq2BF}. Most previous work makes use of attention or copying mechanisms to impose the keyword constraints in a soft manner, which means that the constraint may not be satisfied~\cite{QA,copy}. 

In our CGMH framework, it is natural to impose hard constraints by restricting the support of the stationary distribution to feasible solutions. In particular, we have
 $$\pi(\mathrm x) \propto p_{\text{LM}}(\mathrm x)\cdot \mathcal{X}_{\text{keyword}}(\mathrm x)$$
where $p_{\text{LM}}$ is a general sentence probability computed by a language model and $\mathcal{X}_{\text{keyword}}$ is the indicator function showing if the keywords are included in the generated sentence. In other words, the stationary distribution is proportional to the language model probability if all constraints are satisfied (keywords appearing in the sentence), or 0 otherwise.
During generation, the initial sentence $\mathrm x_0$ is simply a sequence of keywords, and then we perform sampling to generate valid sentences.  
\paragraph{Unsupervised paraphrase generation.}
\label{sec:correction}
Paraphrase generation aims to synthesize literally different sentences that convey the same meaning as the input sentence. 
It is an important task in NLP, and can be a key component in downstream applications such as  data augmentation for NLP. 
Previous state-of-the-art paraphrase generation methods require parallel data for training, which is not always available.
 
In our paper, we view paraphrase generation as sampling from a distribution, where the sentences are (1) fluent by themselves and (2) close in meaning to the input sentence $\mathrm x_*$. The former property can be captured by a traditional language model, whereas the latter can be modeled as a constraint. Concretely, we have
\begin{align}
\label{equ:paraphrase_stationary}
\pi(\mathrm x) \propto p_{\text{LM}}(\mathrm x)\cdot \mathcal X_{\text{match}}(\mathrm x|\mathrm x_*)
\end{align}

Here, $p_{\text{LM}}(\mathrm x)$ is also the probability given by a language model, indicating the fluency of $\mathrm x$. $\mathcal X_{\text{match}}(\mathrm x|\mathrm x_*)$ is a matching score. In our experiments, we have several choices for $\mathcal X_{\text{match}}(\cdot|\cdot)$:
\begin{compactitem}
\item \textit{Keyword matching}~(KW) as hard constraints. We observe that paraphrases typically share some keywords with the original sentence. In this variant, we use Rake~\cite{rose2010automatic} to extract keywords and keep them fixed during sampling. That is $\mathcal X_{\text{match}}(\mathrm x|\mathrm x_*)=1$ if $\mathrm x$ and $\mathrm x_*$ share the same keywords, and 0 otherwise. 

\item \textit{Word embedding similarity} as a soft constraint. Embeddings map discrete words to real-valued vectors, providing a softer way of measuring similarity. In this matching function, we enhance keyword matching with embedding similarity.  For any word $w$ in a sentence $\mathrm x$, we first find the closest word in the input sentence $\mathrm x_*$ by computing their cosine similarity~\cite{pennington2014glove}. Then either the minimum or the average of these cosine measures is taken as the matching score, resulting in two variants~(WVM and WVA).

\item \textit{Skip-thoughts similarity}~(ST) as a soft constraint. The kip-thoughts approach trains sentence embeddings by predicting surrounding sentences~\cite{kiros2015skip}. We compute the cosine similarity between the skip-thought embeddings of $\mathrm x$ and $\mathrm x_*$, and use it as the matching score.
\end{compactitem}
Theoretically speaking, we may start sampling from any sentence, once the stationary distribution is defined.
However, it would take too long for the Markov chain to mix/burn-in (i.e., samples are from the desired distribution). We thus use the original sentence as the initial state, i.e., $\mathrm x_0=\mathrm x_*$. This is similar to the warm start in Gibbs sampling for contrastive divergence estimation~\cite{DBN}.

\paragraph{Unsupervised sentence error correction.}

Previous work of sentence error correction also depends on parallel data~\cite{felice2014grammatical,junczys2016phrase,sakaguchi2017grammatical,chollampatt2016adapting}. Our CGMH framework allows us to generate samples from a distribution of correct sentences, starting from an erroneous one as the input $\mathrm x_*$. In this application, we use the same stationary distribution (Equation \ref{equ:paraphrase_stationary}) as in the unsupervised paraphrase setting, where $p_{\text{LM}}$ is trained on a general corpus ensuring the fluency (correctness), and $\mathcal X_{\text{match}}(\cdot|\cdot)$---assumed insensitive to grammatical errors---imposes a soft constraint of semantic relevance.

\subsection{Acceptance Rate}
In MH, both proposals and stationary distributions can be specified. The way to ensure that the samples are indeed from the desired distribution is to correct the proposal distribution by a probability of acceptance or rejection, given by an acceptance rate in Equations~(\ref{eqn:acceptance1}) and~(\ref{eqn:acceptance2}).

In our approach, we have three types of proposals, namely, deletion, insertion, and replacement. We thus breakdown our acceptance rate (before taking $\min\{1,\cdot\}$) as
\begin{align}\label{eqn:A_replace}
\nonumber A_{\text{replace}}^*(x'|x) 
&=\frac{ p_{\text{replace}}\cdot g_{\text{replace}}(\mathrm x|\mathrm x')\cdot \pi(x')}{p_{\text{replace}}\cdot g_{\text{replace}}(\mathrm x'|\mathrm x)\cdot \pi(x)} \\
&\approx \frac{\pi (w_m|x_{-m})\cdot \pi(\mathrm x')}{\pi (w_m'|x_{-m})\cdot \pi(\mathrm x)}=1\\ \label{eqn:A_insert}
\nonumber A_{\text{insert}}^*(\mathrm x'|\mathrm x)
&=\frac{p_{\text{delete}}\cdot g_{\text{delete}}(\mathrm x|\mathrm x')\cdot \pi(\mathrm x')}{p_{\text{insert}}\cdot g_{\text{insert}}(\mathrm x'|\mathrm x)\cdot \pi(\mathrm x)}\\
&=\frac{p_{\text{delete}}\cdot \pi(\mathrm x')}{p_{\text{insert}}\cdot g_{\text{insert}}(\mathrm x'|\mathrm x)\cdot \pi(\mathrm x)}\\
\label{eqn:A_delete}
\nonumber A_{\text{delete}}^*(\mathrm x'|\mathrm x)&=\frac{p_{\text{insert}}\cdot g_{\text{insert}}(\mathrm x|\mathrm x')\cdot \pi(\mathrm x')}{p_{\text{delete}}\cdot g_{\text{delete}}(\mathrm x'|\mathrm x)\cdot \pi(\mathrm x)}\\
&=\frac{p_{\text{insert}}\cdot g_{\text{insert}}(\mathrm x|\mathrm x')\cdot \pi(\mathrm x')}{p_{\text{delete}}\cdot \pi(\mathrm x)}
\end{align}

In particular, (\ref{eqn:A_replace}) is trivially true because word replacement could be thought of as a step of Gibbs sampling, which is in turn a step of MH sampling whose acceptance rate is guaranteed to be 1. (\ref{eqn:A_insert}) and (\ref{eqn:A_delete}) are reciprocal because deletion and insertion are the inverse operation to each other.

\section{Experiments}
We evaluated our approach on a variety of tasks including sentence generation from keywords, unsupervised paraphrase generation, and unsupervised sentence error correction. 
We also conducted in-depth analysis of the proposed CGMH method.

\subsection{Keywords-to-Sentence Generation}

For keywords-to-sentence generation, we trained our language model on randomly chosen 5M sentences from the {One-Billion-Word Corpus}~\cite{chelba2013one}.\footnote{{http://www.statmt.org/lm-benchmark/}} We held out a 3k-sentence set to provide keywords for testing. 
For each sentence, we randomly sampled one or more words as the constraint(s).
Our language models are simply a two-layer LSTM with a hidden size of 300. We chose 50k most frequent words as the dictionary. 

For MH sampling, we used the sequence of keywords as the initial state, and chose the sentence with the lowest perplexity after 100 steps as the output. We set the maximum sampling step to 200.

We tested the negative likelihood~(NLL) of sentences to evaluate their fluency. 
NLL is given by a third-party $n$-gram language model trained on the English monolingual corpus of WMT18.\footnote{{http://www.statmt.org/wmt18/translation-task.html}}
We also invited 3 volunteers to score the fluency of generated sentences. 
Volunteers were asked to score 100 samples from each method according to their quality. Scores range from 0 to 1, where 1 indicates the best quality. 
\begin{table}[t]\footnotesize
\centering
\resizebox{.8\linewidth}{!}{
\begin{tabular}{|c|c|c|c|c|c|}
\hline
\multicolumn{2}{|c|}{\textbf{\#keyword(s)}}&CGMH&GBS&sep-B/F&asyn-B/F\\\hline
\multirow{2}{*}{1}&NLL&\textbf{7.04}&7.42&7.80&8.30\\\cline{2-6}
 &Human&\textbf{0.45}&0.32&0.11&0.09\\\hline
 \multirow{2}{*}{2}&NLL&\textbf{7.57}&8.72&-&-\\\cline{2-6}
 &Human&\textbf{0.61}&0.55&-&-\\\hline
 \multirow{2}{*}{3}&NLL&\textbf{8.26}&8.59&-&-\\\cline{2-6}
 &Human&\textbf{0.56}&0.49&-&-\\\hline
 \multirow{2}{*}{4}&NLL&\textbf{7.92}&9.63&-&-\\\cline{2-6}
 &Human&\textbf{0.65}&0.55&-&-\\\hline
\end{tabular}
}
\caption{Results of NLL and human evaluation on sentences with 1 to 4 keywords. Sentences with lower NLL and higher human evaluation scores are better.}
\label{tab:res-keyword}
\end{table}

\begin{table}[t]\footnotesize
\centering
\resizebox{.9\linewidth}{!}{
\begin{tabular}{|l|l|}
\hline
\textbf{Keyword(s)}&\textbf{Generated Sentences}\\\hline
friends &My good \textbf{friends} were in danger .\\\hline
project&The first \textbf{project} of the scheme .\\\hline
\multirow{2}{*}{have, trip} &But many people \textbf{have} never \\
&made the \textbf{trip} .\\\hline
\multirow{2}{*}{lottery, scholarships} &But the \textbf{lottery} has provided\\
& \textbf{scholarships} .\\\hline
decision, build, &The \textbf{decision} is to \textbf{build} a new\\
home &\textbf{home} .\\\hline
attempt, copy, &The first \textbf{attempt} to \textbf{copy} the\\
painting, denounced&\textbf{painting} was \textbf{denounced} .
\\\hline	
\end{tabular}}
\caption{Sentences generated from keywords by CGMH. }
\label{tab:example-key}
\end{table}
\begin{table}[!t]\footnotesize
    \centering
    \resizebox{.55\linewidth}{!}{
    \begin{tabular}{|l|c|}
        \hline
        \textbf{Statistic}&\textbf{Value}\\\hline
        Mean intra-annotator std &0.098\\
        Mean intra-model std & 0.280\\ 
        $p$-value~(1 keyword) & $<0.01$\\
        $p$-value~(2 keywords) & $<0.05$\\
        $p$-value~(3 keywords) & $<0.05$\\
        $p$-value~(4 keywords) & $<0.01$\\
        \hline
    \end{tabular}
    }
    \caption{Statistics of human evaluation for keywords-to-sentence generation.}
    \label{tab:human-eval}
\end{table}
Table~\ref{tab:res-keyword} compares our method with current state-of-the-art approaches of constrained generation, namely, the grid beam search approach~(GBS)~\cite{hokamp2017lexically} and two variants of the backward forward model~(sep-B/F and asyn-B/F)~\cite{mou2015backward}.

CGMH outperforms previous work in both NLL and human evaluations.
The two variants of B/F cannot generate more than one keywords. 
GBS is designed for machine translation; it works well when the candidate sentence space is small. 
But for general sentence generation, pruning in grid makes the keywords sometimes unable to find appropriate prefixes. Table~\ref{tab:example-key} provides several examples of keywords-to-sentence generation by \method.

We present statistics of human evaluation in Table~\ref{tab:human-eval}. ``Mean intra-annotator std'' is the mean standard deviation of scores from different volunteers, whereas ``Mean intra-model std'' is the mean standard deviation of each model. This implies that the volunteers achieve high consistency with each other, and that the gap between different models is large. $p$-values in this table are between CGMH and GBS; for \method and B/F models, $p$-values are lower than 0.001. This shows that the results given by human annotation are statistically significant.

\label{sec:exp_para}

\begin{table}[t]
\centering
\resizebox{.9\linewidth}{!}{
\begin{tabular}{|l|c|c|c|}
\hline
\textbf{Model} &\textbf{BLEU-ref} &\textbf{BLEU-ori} &\textbf{NLL} \\\hline\hline
Origin Sentence&30.49&100.00&7.73\\\hline
VAE-SVG~(100k)&22.50&-&-\\
VAE-SVG-eq~(100k)&\textbf{22.90}&-&-\\
VAE-SVG~(50k)&17.10&-&-\\
VAE-SVG-eq~(50k)&17.40&-&-\\
\hline
Seq2seq~(100k) &22.79&33.83&6.37\\
Seq2seq~(50k) &20.18&27.59&\textbf{6.71}\\
Seq2seq~(20k) &16.77&22.44&6.67\\\hline
VAE~(unsupervised) & 9.25& 27.23&7.74 \\\hline
CGMH \textit{w/o matching} & 18.85&50.28 &7.52 \\
\quad \quad \quad \textit{w/ KW} & 20.17 & 53.15 & 7.57 \\
\quad \quad \quad \textit{w/ KW + WVA} & 20.41&53.64 &7.57 \\
\quad \quad \quad \textit{w/ KW + WVM} & 20.89&54.96 &7.46 \\
\quad \quad \quad \textit{w/ KW + ST} & 20.70&54.50 &7.78 \\
\hline
\end{tabular}
}
\caption{Performances of different paraphrase models. 
Ideal paraphrase generator should achieve higher BLEU-ref, lower BLEU-ori, and lower NLL scores.}
\label{tab:para_result}
\end{table}

\begin{table}[!t]\footnotesize
\centering
\resizebox{.9\linewidth}{!}{
\begin{tabular}{|c|l|}
\hline
\textbf{Type}&\textbf{Examples}\\\hline
Ori & what 's the best plan to lose weight \\\hline
Ref & what is a good diet to lose weight \\\hline
Gen & what 's the best way to slim down quickly\\\hline\hline

Ori & how should i control my emotion \\\hline
Ref & how do i control anger and impulsive emotions\\\hline
Gen & how do i control my anger\\\hline\hline

Ori & why do my dogs love to eat tuna fish \\\hline
Ref & why do my dogs love eating tuna fish\\\hline
Gen & why do some dogs like to eat raw tuna and raw fish\\\hline
\end{tabular}}
\caption{Paraphrase generation given by CGMH w/ KW+WVM. For each sample, we show the original sentence (Ori), the reference paraphrase (Ref), and the generated sentence (Gen).}
\label{tab:example-para}

\end{table}

\begin{table}[!t]\footnotesize
\centering
\resizebox{\linewidth}{!}{
\begin{tabular}{|c|l|c|}
\hline
\textbf{Step}&\textbf{State (Sentence)}&\textbf{Proposal}\\\hline
Origin&what movie do you like most .&\textbf{replace} \textit{what} with \textit{which} \\\hline

1&which movie do you like most . &\textbf{delete} \textit{most}\\\hline
2&which movie do you like . &\textbf{insert} \textit{best}\\\hline
3&which movie do you like best . &\textbf{replace} \textit{like} with \textit{think}\\\hline
4&which movie do you think best .  &\textbf{insert} \textit{the}\\\hline
5&which movie do you think the best .  &\textbf{insert} \textit{is}\\\hline
Output&which movie do you think is the best .  &-\\\hline

\end{tabular}}
\caption{An example of the sampling process given by \method w/ KW+WVM.}
\label{tab:r-example-para}

\end{table}

\subsection{Unsupervised Paraphrase Generation}

We followed previous work of supervised paraphrase generation~\cite{gupta2017deep,prakash2016neural,gupta2017deep,li2017paraphrase} and used a standard benchmark, the Quora dataset,\footnote{{https://www.kaggle.com/c/quora-question-pairs/data}} to evaluate each model. 
The dataset contains 140k pairs of paraphrase sentences, and 260k pairs of non-paraphrase sentences. 
We followed the standard dataset split, which holds out 3k and 30k for validation and testing, respectively. 

For supervised baselines, we varied the training samples to be 100k, 50k, and 20k pairs, so that we could evaluate the effect of different parallel data sizes in supervised training.  For unsupervised paraphrase generation, we only need a non-parallel corpus to train the language model. The sentences in the test set, however, are questions, so it is improper to use generic language models (e.g., trained on One-Billion Corpus) to judge the likelihood of a question. Instead, we trained language models on all the training samples that do not appear in the validation and test sets. 
The language models are of the same structure as the ones for keywords-to-sentence generation, except that we reduce the dictionary size to 30k because of fewer training samples. 
 
Previous work uses the BLEU score \cite{papineni2002bleu} against a ground truth reference (denoted as BLEU-ref) to evaluate the quality of the generated paraphrase~\cite{gupta2017deep}. 
We observe that it is insufficient because simply copying the input sentence itself yields the highest BLEU-ref score  (Table~\ref{tab:para_result}). 
We thus propose to use the BLEU score against the original input sentence (denoted as BLEU-orig) as an auxiliary measure. Ideally, BLEU-ref should be high, whereas BLEU-ori should be low.
We tried different initialization states, including using exact or corrupted original sentences. 
We also attempted to start from a totally random state. 
As a lot of samples are generated, we chose the first sentence with BLEU-ori score lower than 55 to compare with other models. (The number is chosen empirically in order to get paraphrases with obvious literal difference.)

We compare our approach with supervised methods, including sequence-to-sequence models, VAE-SVG and VAE-SVG-eq~\cite{gupta2017deep}. VAE-SVG is a VAE conditioned on the original sentence, and VAE-SVG-eq is a variant of VAE-SVG which shares parameters between encoder and decoder.

We would also like to compare paraphrase generator in the unsupervised setting as our \method. However, we cannot find an existing dedicated model. We find it possible to train a variational autoencoder (VAE) with non-parallel corpus and sample sentences from the variational latent space~\cite{K16-1002}.

Table~\ref{tab:para_result} shows, compared with VAE, that our method achieves a fairly close BLEU-ref score to the best supervised approaches. 
Moreover, CGMH even outperforms the supervised methods when the training set is not large enough ($\le$50k). 

Admittedly, CGMH has higher BLEU-ori scores than supervised methods, indicating that the generated samples are closer to the input. {This, however, makes sense because CGMH samples sentences from a distribution specified in an unsupervised fashion, as opposed to rewriting words and phrases in an \textit{ad hoc} fashion to make the expressions different, as is learned in the supervised setting.  
However, we only consider paraphrases with BLEU-ori less than 55, which has assured a significant literal difference.} 
Future research could address this problem by designing proper heuristics to manipulate the stationary distribution, which is beyond the scope of our paper (but shows the flexibility of our model). 

Table~\ref{tab:example-para} shows examples of generated paraphrases. We see qualitatively that CGMH yields fairly good samples in terms of both closeness in meaning and difference in expressions.
Table~\ref{tab:r-example-para} gives a real example of the paraphrase generation  process with CGMH.

\subsection{Unsupervised Error Correction}
\label{sec:correction-exp}

\begin{table}[!t]\footnotesize

\centering

\begin{tabular}{|l|c|c|}
\hline
\textbf{Model}&\textbf{\#parallel data} &\textbf{GLEU}\\\hline
AMU&2.3M&44.85\\
CAMB-14&155k&46.04\\
MLE&720k&52.75\\
NRL&720k&\textbf{53.98} \\\hline
CGMH&0&45.5 \\
\hline
\end{tabular}
\caption{Results of different models on sentence correction.}
\label{tab:result-correction}

\end{table}
\begin{table}[!t]\footnotesize
\centering
\resizebox{\linewidth}{!}{
\begin{tabular}{|c|l|}
\hline
Ori &  Even if \textbf{we are failed} , We have to try to get \textbf{a new things} . \\\hline
Ref & Even if we all failed , we have to try to get new things .\\\hline
Gen & Even if we are failing , We have to try to get some new things .\\\hline\hline

Ori &  In the world \textbf{oil price very high} right now . \\\hline
Ref & In today 's world , oil prices are very high right now .\\\hline
Gen & In the world , oil prices are very high right now .\\\hline

\end{tabular}
}
\caption{Examples of sentence correction by CGMH.}
\label{example-corr}
\end{table}
We evaluated our method on {JFLEG}~\cite{napoles2017jfleg},\footnote{{https://github.com/keisks/jfleg}} a newly released dataset for {sentence correction}. 
It contains 1501 sentences (754 for validation and 747 for test), each with 4 revised references. 
We adopted the GLEU metric \cite{napoles2015ground}, which measures sentence fluency and grammaticality. 

This benchmark dataset does not contain training samples. Various studies have not only proposed new models, but also collected parallel data for themselves, each containing millions of samples, shown in Table~\ref{tab:result-correction}. However, we used none of them.

We adopted the same language models (trained on One-Billion-Word) as in keywords-to-sentence generation to approximate sentence probabilities. To better handle typos and tense errors, we employ en package\footnote{https://www.clips.uantwerpen.be/pages/pattern-en} to provide an additional candidate word set containing possible words with similar spellings or the same root.
For MH sampling, we start from the original sentence, and simply output the 100th sample. 
As the original erroneous sentence has low probability from a language model perspective, the goal of sentence correction can be formulated as jumping to a nearby sentence with high probability. 
We would like to encourage MH to explore more probable states by further rejecting proposals if the likelihood is becoming too small.

The performance of CGMH on error correction is surprisingly promising, as shown in Table~\ref{tab:result-correction}.
CGMH achieves comparable results to CAMB14, a rule-based system for error correction~\cite{felice2014grammatical}. \method even outperforms the AMU system~\cite{junczys2016phrase}, which is built on phrase-based machine translation with 2.3M parallel training pairs and intensively engineered linguistic features. 
We observe some performance gap between CGMH and other supervised approaches, namely, MLE~(Maximal Likelihood Estimation) and NRL~(Neural Reinforcement Learning)~\cite{sakaguchi2017grammatical}. Nevertheless, our initial success of \method shows a promising direction of unsupervised error correction.

\subsection{Model Analysis}
Despite successful applications in the above experiments, we now analyze \method in more detail.
k
\textbf{Acceptance rate and ergodicity.} 
%As is known, it is difficult to quickly traverse the states of a high-dimensional Markov chain, where low acceptance rate is a major problem.  
Table~\ref{tab:accept-rate} shows the acceptance rate in the paraphrase generation task. We see that the word replacement has 100\% acceptance rate as it is essentially a Gibbs step, guaranteed by Equation~\ref{eqn:A_replace}. For word deletion and addition, the acceptance rate is lower, but still in a reasonable range; it allows the sampler to generate sentences with different lengths, as opposed to Gibbs sampling.
In our experiment, it takes about 150 steps to obtain a fluent sentence from a sequence of keywords. 
For paraphrase generation, it takes less than 50 steps for more than 20\% of words being changed, showing that CGMH is efficient for practical use. 

\begin{figure}[!t]
\centering
\includegraphics[width=0.6\linewidth]{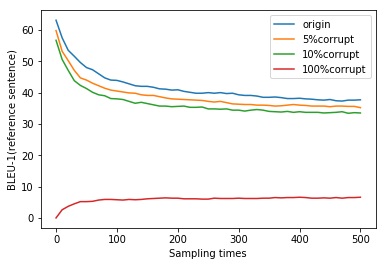}
\caption{\label{fig:par-corrupt}Generation quality with corrupted initial states. At each situation, 0/5\%/10\%/100\% of the words in initial sentences are randomly replaced with other words.}
\end{figure}

\textbf{Initial state of the Markov chain.} 
Theoretically, the initial state of the Markov chain does not affect the stationary distribution, given that the chain is irreducible and aperiodic. Figure~\ref{fig:par-corrupt} shows that, in the experiment of paraphrase, corrupting the initial state $\mathrm x_0$ by a small fraction does not significantly affect model performance.  But if we corrupt 100\% (i.e., start with random initial states), the performance is very low at the beginning but improves gradually. However, it is difficult to obtain satisfactory performance after 100 epochs. This shows that warm start is useful for CGMH sentence sampling.

It should be emphasized that CGMH sampling does not solely reply on the initial state being a valid sentence. In keywords-to-sentence generation, we start from a sequence of keywords, but CGMH eventually yields samples which are generally fluent sentences.

\begin{figure}[!t]
\centering
\includegraphics[width=0.6\linewidth]{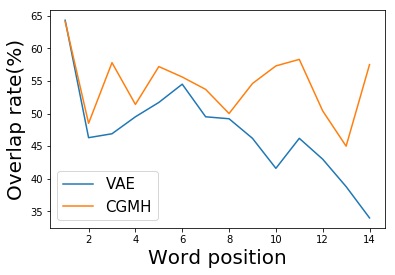}
\caption{\label{fig:overlap}Overlap rates of CGMH and VAE for each word position of sentences.}
\end{figure}

\textbf{Comparison with VAE.}  We would like to compare \method (sampling from the sentence space) with VAE (sampling from the latent space).
VAE is a probabilistic model that imposes a prior distribution on the latent space and then decodes a sentence in a deterministic manner by a RNN. 
In practice, we observe the variance of VAE samples will increase as the generation proceeds. 
This is shown by the blue curve in Figure~\ref{fig:overlap}, as the word overlap rate~(the ratio of the reference sentences containing words at a specific position of the generated ones) goes down for words far from the beginning of a sentence.
This is possibly because RNN can be thought of as an autoregressive Bayesian network generating words conditioned on previous ones. Hence error will accumulate during generation.
However, CGMH does not severely suffer from this problem, because there is not an explicit generation direction~(order) for CGMH. 
At the same time, CGMH has the ability to self-correct, which is shown in the experiment of error correction. 

If we would like to sample diversified sentences, CGMH is also better than VAE, because the diversity is distributed across the entire sentence in CGMH.
\begin{table}[!t]\footnotesize
\centering
\resizebox{.8\linewidth}{!}{
\begin{tabular}{|l|c|c|c|c|}
\hline
\textbf{Model} &\textbf{Rep}&\textbf{Add}&\textbf{Del}&\textbf{Mean}\\\hline
CGMH \textit{w/o matching} &100&9.2&5.1&32.9\\
\quad \quad \quad \textit{w/ KW} &100&8.0&4.4&32.5\\
\quad \quad \quad \textit{w/ KW + WVM} & 100&10.8 & 2.9 & 32.7\\
\hline
\end{tabular}}
\caption{Acceptance rate (\%) in the paraphrase generation task. Word replacement has 100\% acceptance rate as shown in Equation~(\ref{eqn:A_insert}).}
\label{tab:accept-rate}  
\end{table}

\section{Conclusion}

In this paper, we present \method for constrained sentence generation by Metropolis-Hastings (MH) sampling, where we propose word-level operations including replacement, insertion, and deletion as proposals, and design several stationary distributions for different tasks. 
We evaluated our results on keywords-to-sentence generation, paraphrase generation, and error correction.
Our CGMH framework not only makes unsupervised learning feasible in these applications, but also achieves high performance close to state-of-the-art supervised approaches. 

\section{Acknowledgments}
We would like to thank the anonymous reviewers for their insightful comments.
This work is supported by the National Key Research and Development Program of China (No.~2017YFC0804001) and the National Science Foundation of China (Nos.~61876196 and~61672058).
\bibliography{aaai19}
\bibliographystyle{aaai}

\end{document}